\documentclass[11pt]{article}

\usepackage[final]{acl}

\usepackage{times}
\usepackage{latexsym}
\usepackage{paralist}
\usepackage{booktabs}
\usepackage{amssymb} 
\usepackage{hyperref}
\usepackage{subcaption}

\usepackage[T1]{fontenc}

\usepackage[utf8]{inputenc}

\usepackage{microtype}
\usepackage{amsmath}

\usepackage{inconsolata}

\usepackage{graphicx}
\usepackage{enumitem}
\usepackage{booktabs}  
\usepackage{multirow}  

\usepackage[most]{tcolorbox}
\usepackage{xcolor}
\usepackage{enumitem}
\usepackage{fontawesome5} 

\definecolor{badprompt}{RGB}{255, 240, 240} 
\definecolor{badborder}{RGB}{200, 100, 100}
\definecolor{goodprompt}{RGB}{240, 248, 255} 
\definecolor{goodborder}{RGB}{70, 130, 180}
\definecolor{analysisbg}{RGB}{245, 245, 245}

\newtcolorbox{promptbox}[2][]{
  enhanced,
  breakable,
  title={#2},
  colback=white,
  colframe=black!70,
  fonttitle=\bfseries,
  colbacktitle=black!10,
  coltitle=black,
  attach boxed title to top left={yshift=-2mm, xshift=2mm},
  boxrule=0.5pt,
  #1
}

\newcommand{\originalprompt}[1]{
    \begin{tcolorbox}[
        colback=badprompt,
        colframe=badborder,
        title=\textbf{\small \faTimesCircle\ Original Prompt (Fragile)},
        fonttitle=\bfseries\small,
        coltitle=badborder!60!black,
        colbacktitle=badprompt,
        boxrule=0.5pt,
        arc=2mm,
        left=2pt, right=2pt, top=2pt, bottom=2pt
    ]
    \small \texttt{#1}
    \end{tcolorbox}
}

\newcommand{\improvedprompt}[1]{
    \begin{tcolorbox}[
        colback=goodprompt,
        colframe=goodborder,
        title=\textbf{\small \faCheckCircle\ Improved Prompt (Robust)},
        fonttitle=\bfseries\small,
        coltitle=goodborder!60!black,
        colbacktitle=goodprompt,
        boxrule=0.5pt,
        arc=2mm,
        left=2pt, right=2pt, top=2pt, bottom=2pt
    ]
    \small \texttt{#1}
    \end{tcolorbox}
}

\title{Beyond Prompt Engineering: A Systematic Analysis of Prompt Lexical Sensitivity and Its Impacts on Quality}

\author{
  \textbf{Qipeng Xie\textsuperscript{1,5}}\thanks{Equal contributions.},
  \textbf{Zi Liang\textsuperscript{2}}\footnotemark[1],
  \textbf{Jiafei Wu\textsuperscript{3}}\footnotemark[2],
  \textbf{Yufei Chen\textsuperscript{3}},
  \textbf{Weizheng Wang\textsuperscript{2}},
  \\
  \textbf{Wenao Ma\textsuperscript{4}}\footnotemark[2],
  \textbf{Zhong Ming\textsuperscript{1}},
  \textbf{Haiqin Yang\textsuperscript{1}}\thanks{Corresponding authors.},
  \textbf{Kaishun Wu\textsuperscript{5}}
  \\
  \textsuperscript{1}Shenzhen Technology University,
  \textsuperscript{2}The Hong Kong Polytechnic University,
  \textsuperscript{3}Zhejiang Lab
  \\
  \textsuperscript{4}The Chinese University of Hong Kong,
  \textsuperscript{5}HKUST (Guangzhou)
  \\
  \vspace{0.1cm}
  \small
  \textsuperscript{1}\{qxieaf@connect.ust.hk, wuks@hkust-gz.edu.cn\}
  \textsuperscript{2}\{zi1415926.liang@connect.polyu.hk, weizheng.wang@ieee.org\}
  \\
  \small
  \textsuperscript{3}\{wujiafei@zhejianglab.org, cyf200409@gmail.com\}
  \textsuperscript{4}wenaoma@gmail.com
  \textsuperscript{5}\{yanghaiqin@sztu.edu.cn\}
}

\begin{document}
\maketitle
\begin{abstract}
Large Language Models (LLMs) exhibit extreme sensitivity to surface-level prompt variations, in which minor lexical changes can trigger disproportionate performance fluctuations.  Moving beyond black-box optimization and coarse-grained templates, we present the first large-scale, n-gram token-level mechanistic analysis of prompt stability, leveraging a dataset of 132,000 prompt variants.  Our investigation reveals a fundamental Scaling Law of Prompt Performance Stability: higher average task performance is strongly associated with lower variance and greater robustness across prompt perturbation.  We identify two core linguistic drivers underlying this robustness: (1) Domain-Specific Terminology, which tightly anchors semantic boundaries, and (2) Explicit Action Directives, which formalize reasoning trajectories.  Together, these elements constrain the model's interpretative space, effectively ``locking in'' more deterministic generation behavior.  Building on these insights, we introduce an automated Prompt-Refining Agent that systematically restructures input queries by injecting domain anchoring and operational constraints.  Empirical evaluation shows that our approach reduces performance variance by 40.7\% in code generation task, while preserving or improving mean performance.  These findings provide a statistically grounded and mechanistically interpretable framework for achieving robust prompt engineering.
\end{abstract}

\section{Introduction}
\begin{figure}
    \centering
    \includegraphics[width=\linewidth]{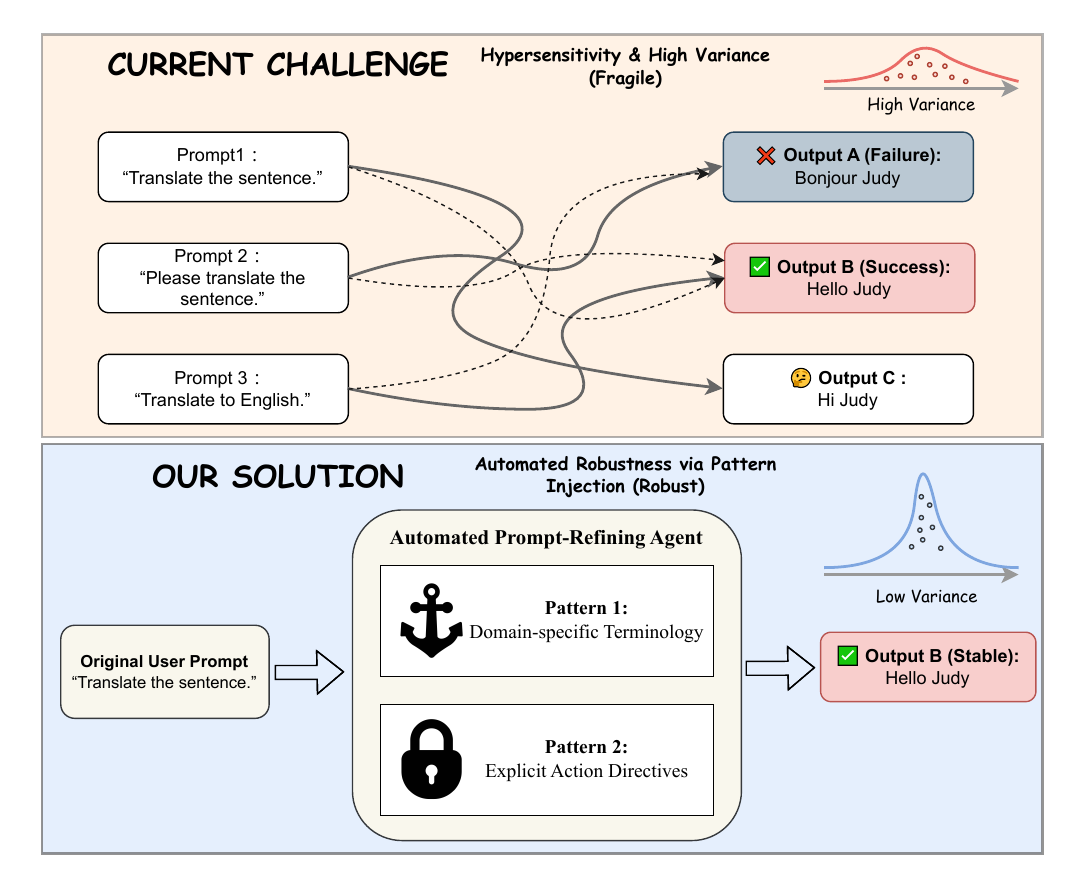}
    \caption{\textbf{The Prompt Sensitivity Challenge vs. Our Solution.} Top: Minor lexical variations in prompts trigger drastic performance fluctuations (high variance). Bottom: Our Automated Prompt-Refining Agent stabilizes generation by injecting domain-specific terminology and explicit action directives (low variance).}
    \label{fig:intro}
\end{figure}

With the rapid advancement of large language models (LLMs), systematically evaluating their robustness and consistency across different prompt formulations has become increasingly critical~\cite{liu2023pre,dong2024survey,mei2025surveycontextengineeringlarge}. As illustrated in Fig.~\ref{fig:intro} (top), prompt engineering often suffers from a ``butterfly effect'': even minor lexical variations can lead to substantial performance fluctuations on the same task~\cite{lu2024prompts}.

This phenomenon, termed prompt instability, undermines the reliability and reproducibility required for trustworthy LLM deployment.  While prompt engineering has emerged as a practical solution to alleviate this issue, it remains largely heuristic and reliant on trial-and-error, lacking a principled understanding of the underlying mechanisms.  Consequently, a fundamental question persists: \textit{\textbf{What determines whether a prompt achieves both high average performance and stable behavior under stylistic perturbations?}}

Existing research generally follows two main directions.  The first focuses on performance optimization through techniques such as calibration~\cite{zhao2021calibrate}, ensembling~\cite{sclar2024quantifying}, regularized training~\cite{liu2021gpt}, and format mixing~\cite{perez2021truefewshotlearninglanguage}.  Although often empirically effective, these methods primarily target performance gains and provide limited insight into the mechanisms governing prompt effectiveness and robustness. The second line of work conducts template-level~\cite{zhu2024promptbenchunifiedlibraryevaluation,pezeshkpour2024large} or decoding-level~\cite{sclar2024quantifying} analyses.  However, these studies typically operate at a coarse granularity, making it difficult to pinpoint how fine-grained token-level patterns influence model robustness.  Moreover, existing sensitivity analyses are often either overly abstract or rely on gradient-based saliency methods~\cite{mizrahi2024stateartmultipromptllm} applied to small-scale samples, limiting their statistical power and generalizability.

To address this gap, we adopt a complementary, mechanism-oriented perspective.  Instead of proposing new optimization techniques, we conduct a systematic analysis of prompt sensitivity at the \textbf{n-gram token level}, an unprecedented scale of grainularity, through large-scale statistical examination.  Unlike recent gradient-based approaches limited to individual examples~\cite{mizrahi2024stateartmultipromptllm}, we analyze 12,000 instructions, each with 11 stylistic variants (132,000 evaluations in total).  This enables us to uncover interpretable token-level patterns that characterize high-robustness prompts.  Furthermore, we reveal a fundamental scaling law linking a prompt's average performance to its stability (see Fig.~\ref{fig:intro}, bottom), offering a new mechanistic lens for prompt engineering. 

Our key contributions are as follows:
\begin{itemize}
    \item We present \textbf{the first large-scale, n-gram–level statistical analysis of prompt sensitivity}, based on a comprehensive dataset of 132,000 prompt variations.  This provides fine-grained, statistically robust insights into prompt behavior.
    \item We empirically discover and validate \textbf{a scaling law between prompt performance and robustness}, demonstrating that prompts with higher average performance consistently exhibit greater stability under stylistic perturbations.
    \item We identify \textbf{two key linguistic patterns}, domain-specific terminology and explicit action directives, that underpin high-robustness prompts.  Building on these insights, we develop an automated \textbf{Prompt-Refining Agent} that constrains the model’s interpretative space, achieving a 40.7\% reduction in performance variance while maintaining or improving mean performance.
\end{itemize}

\section{Related Work}
\paragraph{Prompt Engineering \& Sensitivity Analysis.}
Prompt engineering aims to unlock LLM capabilities through carefully crafted input.  However, research consistently demonstrates that LLMs exhibit hypersensitivity to surface-level prompt variations: even semantically equivalent perturbations can trigger drastic performance volatility.  The ubiquity of this phenomenon is systematically quantified, identifying model susceptibility to spurious feature dependence and formatting shifts as primary drivers~\cite{sclar2024quantifying,voronov-etal-2024-mind}.
This sensitivity extends beyond lexical phrasing to contextual structure.  Empirical evidence~\cite{pezeshkpour2024large} reveals that the permutation of answer options in multiple-choice tasks significantly biases decision-making.  Furthermore, in few-shot learning contexts~\cite{lu2022fantastically}, the ordering of demonstrations often outweighs the impact of the example count itself.
To quantify such sensitivity, recent works~\cite{zhuo2024prosa,lu2024prompts} propose metric-based frameworks. For instance, the PROSA~\cite{zhuo2024prosa} is introduced to assess robustness under prompt perturbations, while others~\cite{lu2024prompts} utilize gradient-based saliency analysis to characterize sensitivity differences. Nevertheless, these approaches predominantly operate at the coarse-grained template level, lacking a systematic statistical attribution of fine-grained lexical units across large-scale instruction datasets.

\paragraph{Prompt Optimization \& Calibration.}
To mitigate instability caused by prompt sensitivity, prior studies~\cite{zhao2021calibrate,zhou2024batch,yao2024prompt} investigate inference-time calibration and training-time consistency constraints.  During inference, the ``calibrate-before-use''~\cite{zhao2021calibrate} minimizes inter-prompt variance by recalibrating output distributions, and another method~\cite{zhou2024batch} extends to batch calibration for in-context learning. During training, \citet{yao2024prompt} introduces Prompt Perturbation Consistency Learning, which regularizes models to maintain output consistency across paraphrased inputs.
Different from those that treat prompt sensitivity as noise to be eliminated, this work posits that sensitivity is a structured statistical phenomenon. Through large-scale, fine-grained analysis, we aim to uncover the intrinsic linguistic properties of high-performing prompts, thereby explaining the mechanism behind their inherent robustness.

\section{Methodology}
This section details a systematic prompt perturbation framework based on orthogonal rewriting strategies.  Unlike prior work relying on random noise or single paraphrases, the proposed design enables multidimensional stress testing of prompts under realistic user input variations.

\subsection{Systematic Perturbation Strategies}
\label{sec:5strategies}
To simulate the diversity of real-world user inputs and examine model sensitivity across multiple dimensions, we define five semantically equivalent yet stylistically distinct rewriting strategies.  For each original instruction, the Gemini-2.5-flash model \cite{comanici2025gemini} is used to generate a corresponding variant under each strategy, as detailed in Appendix~\ref{sec:rewrite_prompt}. 

The definitions of the five strategies are as follows:
\textbf{(1) Semantic Equivalence Paraphrasing (SEP):} Involving extensive syntactic reorganization and synonym substitution, this strategy strictly preserves core task semantics while simulating the natural variations in how different users articulate the same intent~\cite{zhu2024promptrobustevaluatingrobustnesslarge, moore2025evaluatingautoformalizationrobustnesssemantically}.
\textbf{(2) Adding Contextual Information (ACI):} By injecting substantial unrelated background information—such as historical trivia or fiction—we challenge the model's attention stability, testing its ability to extract core directives from noisy, long-tail contexts~\cite{shi2023largelanguagemodelseasily, yoran2024makingretrievalaugmentedlanguagemodels}.
\textbf{(3) Changing Format/Style (CFS):} The instruction's tone is transformed into distinct pragmatic registers (\textit{e.g.}, legal, academic, or casual) to evaluate the model's robustness against significant shifts in linguistic style and formality~\cite{ngweta2025llmsrobustnesschangesprompt}.
\textbf{(4) Introducing Ambiguity (IA):} Uncertainty is introduced through hedging terms (\textit{e.g.}, ``roughly'') and conditional clauses, assessing the stability of the model's reasoning processes when facing non-rigid or probabilistic constraints~\cite{kim2024aligninglanguagemodelsexplicitly}.
\textbf{(5) Syntactic Noise (SN):} Simulating low-quality user inputs, this strategy perturbs the surface form with orthographic errors, including typos and irregular spacing, while maintaining basic intelligibility at the character level~\cite{liu2025evaluatingrobustnesslargelanguage}.

\subsection{Methodological Validation}
We validate the proposed strategy set as a measurement tool by examining its orthogonality in embedding space and its coverage of linguistic features.

\subsubsection{Orthogonality of Perturbation Directions}
It is essential to verify that different strategies probe distinct dimensions of model capability rather than redundantly testing the same feature. To this end, for each rewritten instruction, we compute the delta vector $\Delta \mathbf{v}$ in the embedding space.
\begin{equation}
\Delta \mathbf{v} = \mathbf{v}_{\text{rewrite}} - \mathbf{v}_{\text{original}},
\end{equation}
where $\mathbf{v(\cdot)}$ denotes the embedding function used in our experiments.

\begin{figure}
    \centering
    \includegraphics[width=\linewidth]{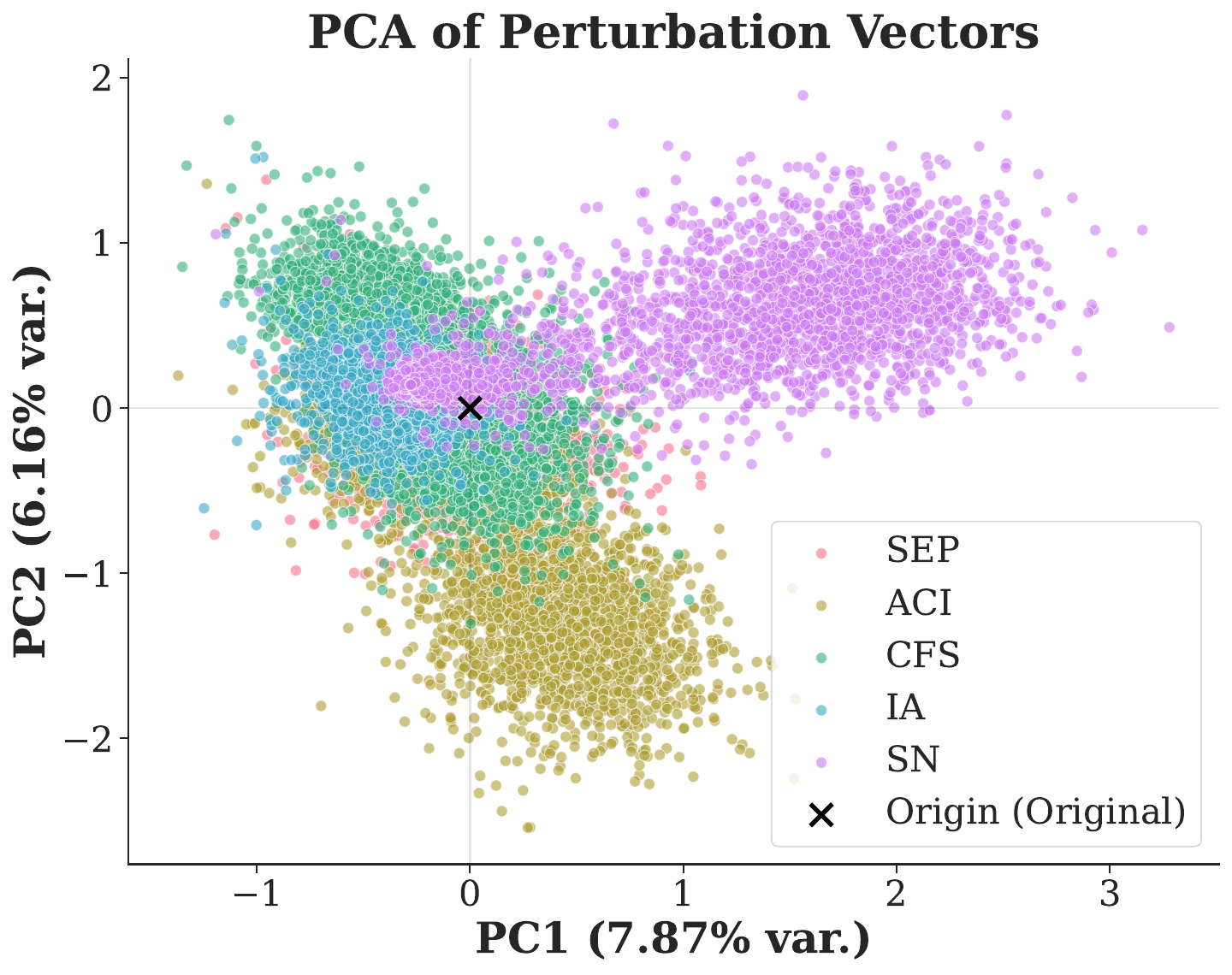}
    \caption{PCA visualization of delta vectors. Each point represents the embedding displacement
induced by one rewritten prompt; the origin corresponds to zero displacement.}
    \label{fig:PCA}
\end{figure}

\begin{figure*}
    \centering
    \includegraphics[width = 16cm]{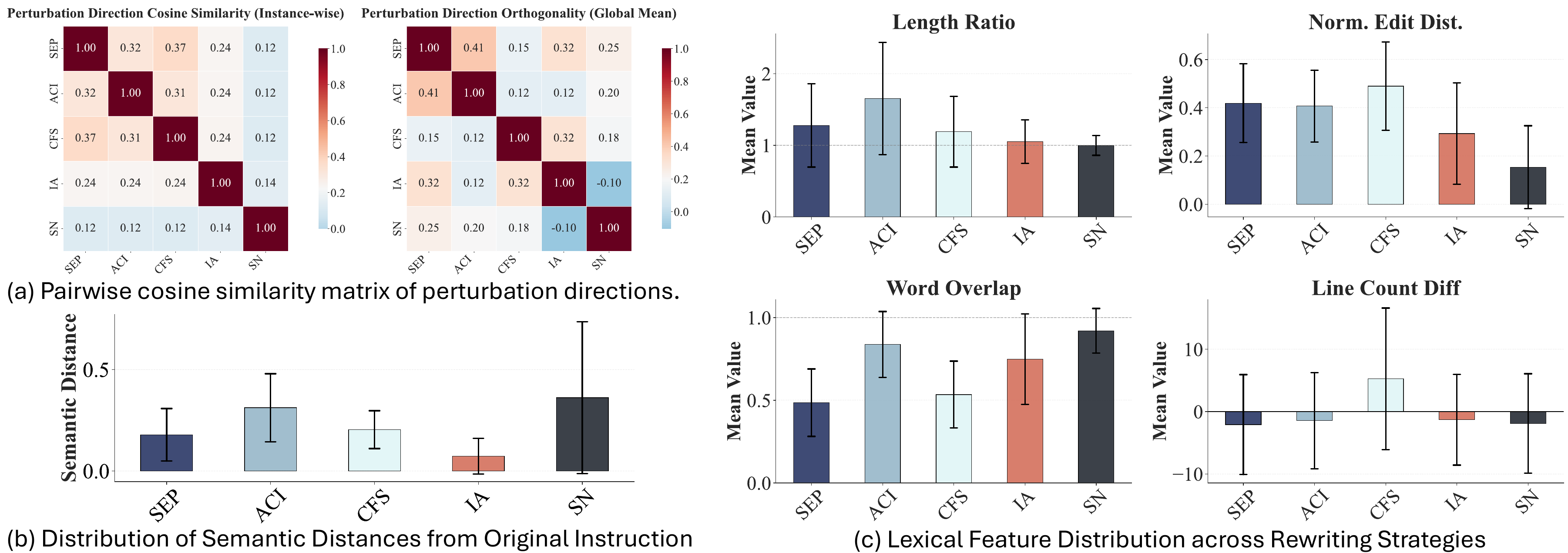}
    \caption{(a) Pairwise cosine similarity matrix of perturbation directions. (b) Average semantic distance from the original instruction by strategy, illustrating a graded perturbation spectrum. (c) Lexical feature analysis, showing that strategies induce measurable variation in surface form and structure.}
    \label{fig:similar}
\end{figure*}

\paragraph{Cluster Analysis} Principal Component Analysis (PCA) \cite{mackiewicz1993principal} on the $\Delta \mathbf{v}$ reveals that samples from different strategies from distinct, well-separated clusters in the project space.  As shown in Figure \ref{fig:PCA}, SN and ACI exhibit large displacements from the origin, whereas IA remains clustered near the center. 
\paragraph{Cosine Similarity} We compute pairwise cosine similarity between $\Delta \mathbf{v}$ of different strategies. The results shown in Figure \ref{fig:similar} (a) indicate a low average off-diagonal similarity of approximately 0.22. Notably, the similarity between SN and others is even lower ($\approx 0.12$).These findings demonstrate that the five strategies induce highly independent perturbation directions, minimizing redundancy in the evaluation.

\subsubsection{Coverage of Perturbation Intensity}
Beyond semantic vector analysis, we analyze lexical features to verify the coverage of non-semantic dimensions.
\paragraph{Literal and Informational Coverage} Lexical feature analysis in Fig.~\ref{fig:similar} (c) demonstrates that SN yields a high normalized edit distance, effectively covering literal and surface-form perturbations. Meanwhile, ACI results in a length ratio significantly greater than 1.0, achieving systematic information expansion.
\paragraph{Intensity Spectrum} We further compare strategies by their average semantic distance to the original instruction. Figure \ref{fig:similar} (b)
shows that the five strategies span a range of perturbation intensities. This ranges from the micro-perturbations of IA (average distance $\approx 0.07$), transitioning through the mid-range of SEP and CFS, to the strong perturbations of SN (average distance $\approx 0.36$).
\section{Experiments}
This section validates the systematic perturbation framework via large-scale empirical studies, quantifies the relationship between prompt performance and robustness, and dissects the underlying mechanisms from a microscopic perspective.
\subsection{Settings}
\paragraph{Dataset}
To ensure comprehensive coverage of diverse instruction types and complexity levels, we employ the WizardLM\_evol\_instruct\_70k \cite{luo2025wizardmathempoweringmathematicalreasoning} dataset, which consists of highly complex instructions generated through multiple rounds of evolutionary optimization, including tasks such as code writing, creative writing, and math calculation. From this dataset, we randomly sample 12,000 instances to form the core experimental benchmark, striking a balance between statistical significance and computational cost.

\paragraph{Pipeline}
Our experimental framework employs a three-model pipeline.  The \textbf{Rewriter} utilizes Gemini-2.5-flash~\cite{comanici2025gemini} to generate 10 variants for each original instruction based on the five strategies defined in Sec.~\ref{sec:5strategies}, resulting in a set of 11 prompts per instance (1 original and 10 variants).  The \textbf{Generator} employs \texttt{Qwen-plus}~\footnote{\href{https://modelstudio.console.alibabacloud.com/?tab=doc\#/doc/?type=model\&url=2840914_2\&modelId=qwen-plus}{\nolinkurl{modelstudio.console.alibabacloud.com}}} \cite{yang2025qwen3} to synthesize responses for the entire corpus of 132,000 instructions ($12k \times 11$). Finally, to strike a balance among model intelligence, inference speed, and cost efficiency, we employ \texttt{Grok-4-fast}~\footnote{\url{https://x.ai/news/grok-4-fast}} as \textbf{Evaluator} to perform fully automated quality assessments adopting the LLM-as-a-Judge paradigm~\cite{zheng2023judging, liu2023geval, gu2025surveyllmasajudge}.

\paragraph{Evaluation Metrics}
To capture fine-grained quality nuances, we implement a continuous 1-100 scoring mechanism (Appendix~\ref{sec:eval_prompt}) that explicitly counters the central tendency bias prevalent in LLM judges~\cite{wang2023largelanguagemodelsfair, zheng2023judging}. Adopting a `Utility Hierarchy' aligned with HELM~\cite{liang2023holisticevaluationlanguagemodels} and FLASK~\cite{ye2023flask}, we prioritize foundational validity before addressing robustness.  Specifically, we introduce a composite metric $\mathcal{S}$ based on a \textit{utility hierarchy}. As shown in Eq.~(\ref{eq:score_metric}), the weighting scheme prioritizes \textit{foundational validity} over stylistic features: \textbf{Accuracy} ($0.30$) and \textbf{Relevance} ($0.25$) are assigned the highest weights (cumulatively $0.55$) to strictly penalize hallucinations and off-topic responses. \textbf{Completeness} ($0.20$) and \textbf{Clarity} ($0.15$) follow to ensure constraint adherence and readability, while \textbf{Practical Value} ($0.10$) serves as a supplementary metric to reward actionable insights:
\begin{align}
\mathcal{S} &= 0.30\cdot\mathrm{Accuracy}+0.25\cdot\mathrm{Relevance} \nonumber \\
&+0.20\cdot\mathrm{Completeness}+0.15\cdot\mathrm{Clarity} \nonumber \\
&+0.10\cdot\mathrm{Practical\ Value},
\label{eq:score_metric}
\end{align}

Then, we compute the Mean Score ($\mu$) and Standard Deviation ($\sigma$) across each group of 11 samples, serving as the core metrics for quantifying prompt performance and sensitivity based on the LLM-as-a-Judge.

\subsection{The Scaling Law of Prompt Performance Stability}
\begin{figure}[t]
    \centering
    \includegraphics[width=\linewidth]{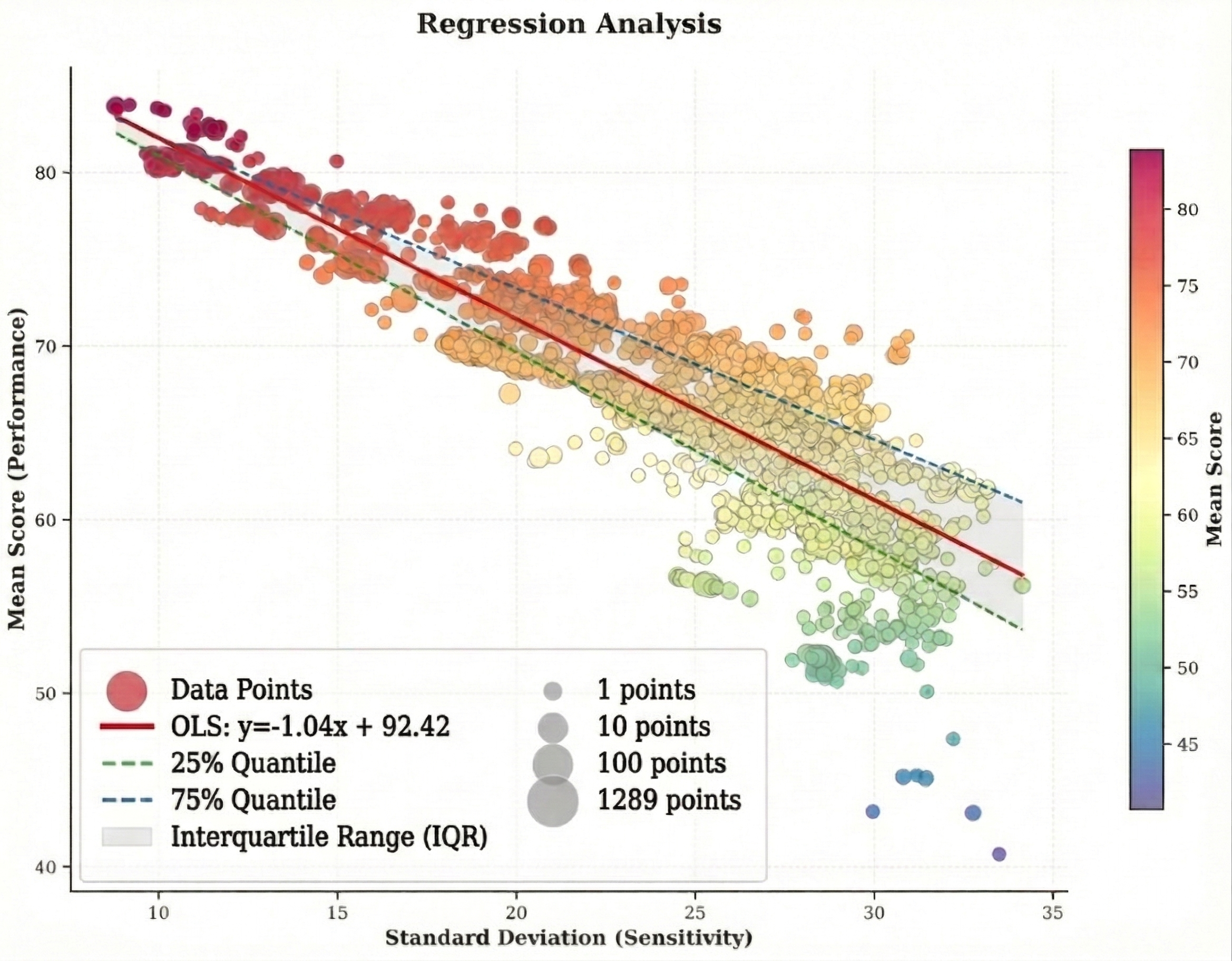}
    \caption{\textbf{The Scaling Law of Prompt Performance Stability.} Scatter plot showing mean performance vs. standard deviation across 12,000 instructions with 11 variants each (132,000 total evaluations). The red line shows OLS regression ($y = -1.083x + 93.60$, $R^2 = 0.7006$). Color gradient indicates point density.}
    \label{fig:scaling_law}
\end{figure}

Figure~\ref{fig:scaling_law} reveals a striking negative correlation between mean performance and response variability.  Through Ordinary Least Squares (OLS) regression, we obtain:
\begin{equation}
y = -1.083x + 93.60, \quad R^2 = 0.7006
\end{equation}
where $y$ represents the mean performance score and $x$ denotes the standard deviation (prompt performance stability). 

\textbf{This relationship establishes a fundamental scaling law}: prompts with higher average task performance exhibit systematically greater robustness to stylistic variations.  The estimated negative slope of $-1.083$ quantitatively characterizes this trade-off, indicating that a one-unit increase in performance variability (standard deviation) is associated with an average decrease of approximately 1.08 points in mean performance. Moreover, the coefficient of determination ($R^2 = 0.70$) suggests that robustness to stylistic perturbations accounts for a substantial proportion (70\%) of the observed variance in prompt performance.

\subsubsection{Implications}
The identified scaling law carries profound implications across three critical dimensions: \textbf{(1) Methodologically}, it necessitates re-conceptualizing prompt sensitivity—not as transient measurement noise to be averaged out, but as a fundamental dimension of model capability that warrants explicit reporting. \textbf{(2) Practically}, it guides production deployment strategies to prioritize prompt designs situated on the \textit{high-performance, low-variability} frontier of this scaling relationship; and \textbf{(3) Mechanistically}, the intrinsic correlation suggests that robustness and performance stem from shared underlying linguistic factors, a hypothesis we dissect in the subsequent section.


\section{Token-Level Analysis}
In this section, we further investigate token-level factors that contribute to prompt robustness and design a controlled experiment to examine whether explicitly injecting these patterns into standard prompts can effectively enhance robustness across diverse tasks.
\begin{table*}[htbp]
\centering
\resizebox{0.9\textwidth}{!}{%
\begin{tabular}{lccc}
\toprule
\textbf{Quadrant} & \textbf{Criteria} & \textbf{Behavioral Archetype} & \textbf{Utility Assessment} \\
\midrule
\textbf{HP-HR} & $\mu > \text{Med}, \sigma < \text{Med}$ & \textit{The Ideal State} & \textbf{Optimal}: Consistently high quality; resilient to perturbations. \\
\addlinespace
HP-LR & $\mu > \text{Med}, \sigma > \text{Med}$ & \textit{High Risk, High Return} & \textbf{Unreliable}: High potential but statistically unstable output. \\
\addlinespace
LP-HR & $\mu < \text{Med}, \sigma < \text{Med}$ & \textit{Consistent Mediocrity} & \textbf{Limited}: Stable behavior but persistently fails to meet quality standards. \\
\addlinespace
LP-LR & $\mu < \text{Med}, \sigma > \text{Med}$ & \textit{Chaotic Failure} & \textbf{Detrimental}: Both low-quality and highly unpredictable. \\
\bottomrule
\end{tabular}%
}
\caption{\textbf{Quadrant Stratification Taxonomy.} We partition prompts into four distinct categories based on median thresholds of $\mu$ and $\sigma$. The \textbf{HP-HR} group represents the ideal target for robust prompt engineering.}
\label{tab:quadrants}
\end{table*}
\subsection{Stratification Strategy}
To disentangle token-level linguistic mechanisms of robustness, we map the 12,000 core instructions onto a ``Performance-Sensitivity'' coordinate system. By applying the median Mean Score ($\mu$) and median Standard Deviation ($\sigma$) as orthogonal thresholds, we partition the dataset into four mutually exclusive quadrants, see Table \ref{tab:quadrants} for details. This strategy specifically isolates the High-Performance, High-Robustness (HP-HR) group, which we designate as the ``Ideal State'' of prompt engineering for its ability to simultaneously achieve high quality and stability. We then conduct a contrastive extraction of distinctive n-grams ($n=1,2,3$) using Log Odds Ratio with add-$k$ smoothing ($k=0.01$)\cite{monroe2008fightin}. This statistical approach rigorously identifies tokens that are \textit{statistically overrepresented} in robust prompts compared to the general population.


\begin{figure*}[htbp]
    \centering
    \includegraphics[width=1\linewidth]{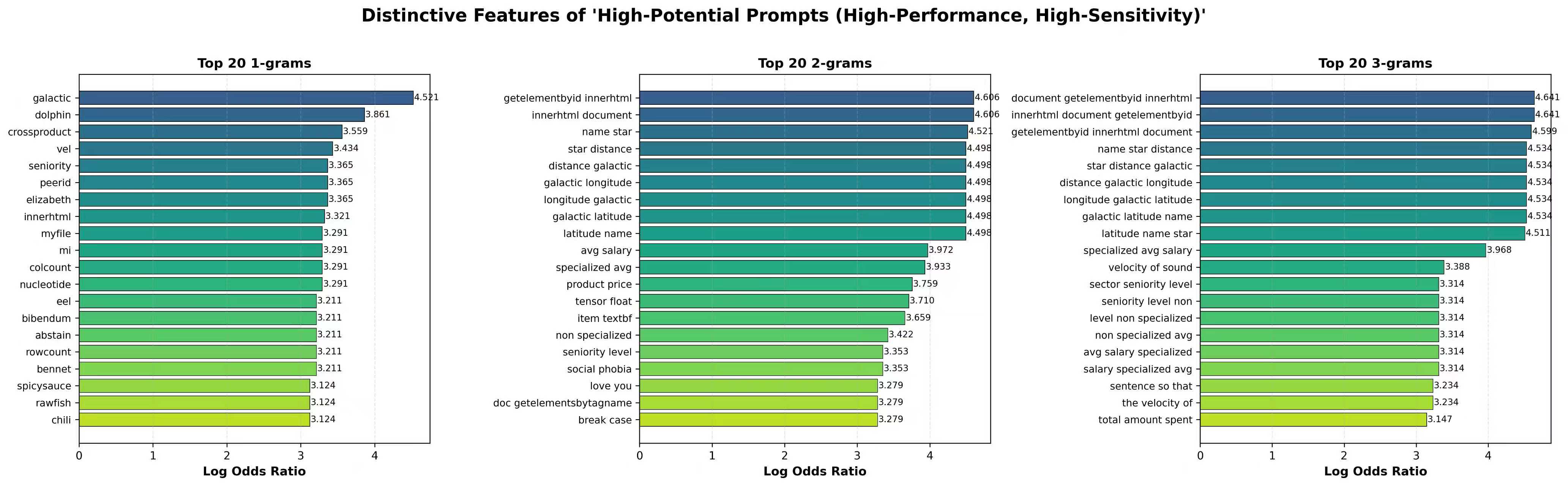}
    \caption{\textbf{Distinctive Token Patterns in High-Robustness Prompts.} Top-20 n-grams ranked by Log Odds Ratio for the high-performance, high-robustness quadrant (N=3,247 instructions) compared to the union of other quadrants (N=8,753). Left: 1-grams (domain terminology). Center: 2-grams (action phrases). Right: 3-grams (structured directives). Error bars show 95\% confidence intervals via bootstrap resampling (1,000 iterations).}
    \label{fig:token_patterns}
\end{figure*}

\subsection{Key Findings: Interpretable Token Patterns}\label{sec:keyfindings}
Figure~\ref{fig:token_patterns} presents the top-20 most distinctive n-grams for the HP-HR quadrant. Our analysis reveals two prominent patterns: Domain-Specific Terminology and Explicit Action Directives.

\paragraph{Pattern 1: Domain-Specific Terminology (Semantic Anchors)} Distinctive 1-grams are dominated by specialized terms (e.g., ``algorithm'', ``protocol'', ``specification'')  that drastically narrow the semantic search space.

\noindent\textbf{Example (Code Writing):}
\begin{itemize}
    \setlength\itemsep{0em} 
    \item[$\times$] \textbf{Vague:} ``Check the \textbf{inputs}'' $\rightarrow$ Ambiguous (data types vs.\ logical validity?), leading to high variance.
    \item[$\checkmark$] \textbf{Robust:} ``Validate \textbf{function parameters}'' $\rightarrow$ Locks the model into a standard programming context, acting as a semantic anchor.
\end{itemize}

\paragraph{Pattern 2: Explicit Action Directives (Operational Anchors)} While domain terms define what to discuss, distinctive n-grams like ``compare and contrast'' and ``step-by-step explanation'' define how to structure the response.

\noindent\textbf{Example (Analytical Reasoning):}
\begin{itemize}
    \setlength\itemsep{0em}
    \item[$\times$] \textbf{Open-ended:} ``Tell me about X'' $\rightarrow$ High structural degrees of freedom.
    \item[$\checkmark$] \textbf{Robust:} ``\textbf{Compare and contrast} X using \textbf{bullet points}'' $\rightarrow$ Constrains the execution path and output scaffolding, eliminating the ``butterfly effect'' of lexical perturbations.
\end{itemize}
High-potential prompts thus enhance robustness by minimizing the interpretation space. simultaneously specifying the semantic domain and the structural template.

\subsection{Automated Optimization via Prompt-Refining Agent}
To validate the causal utility of our findings and provide a scalable solution, we design an automated Prompt-Refining Agent. Instead of treating prompt engineering as a manual heuristic, this agent autonomously transforms fragile user queries into robust instructions by strictly adhering to the Pattern Injection principles identified in Sec.~\ref{sec:keyfindings}.
\subsubsection{Pattern Injection}
The agent functions through a two-stage injection pipeline, explicitly designed to minimize the interpretation space:
\begin{itemize}
    \item Domain Anchoring: Replacing generic descriptions with Domain-Specific Terminology to narrow the semantic search space (e.g., refining ``fix the code'' into ``debug the implementation based on syntax standards'').
    \item Constraint Refinement: Injecting clear formatting and operational directives to lock the execution path (e.g., appending ``provide a step-by-step derivation'' or ``format as bullet points''), the agent injects Explicit Action Directives.
\end{itemize}
We then conduct a comparative evaluation (Base vs. Improved) across 2,000 samples per task.  Our primary metric for robustness is the change in Standard Deviation (Std), while Mean Score (Mean) serves as the indicator for performance quality.

\begin{table*}[t]
\centering
\label{tab:main_results}
\resizebox{0.95\textwidth}{!}{%
\begin{tabular}{llcccccc}
\toprule
\multirow{2}{*}{\textbf{Task Domain}} & \multirow{2}{*}{\textbf{Metric}} & \multicolumn{2}{c}{\textbf{Base Prompt}} & \multicolumn{2}{c}{\textbf{Improved Prompt (Ours)}} & \multicolumn{2}{c}{\textbf{Improvement}} \\
\cmidrule(lr){3-4} \cmidrule(lr){5-6} \cmidrule(lr){7-8}
 & & Mean ($\uparrow$) & Std ($\downarrow$) & Mean ($\uparrow$) & Std ($\downarrow$) & $\Delta$ Mean & $\Delta$ Std \\
\midrule

\multirow{2}{*}{Code Writing} & Rouge-L & 16.77 & 3.24 & \textbf{17.40} & \textbf{1.92} & +3.8\% & -40.7\% \\
 & LLM-as-a-Judge & 77.17 & 12.96 & \textbf{83.22} & \textbf{7.73} & +7.9\% & -40.4\% \\
\midrule 

\multirow{2}{*}{Creative Writing} & Rouge-L & \textbf{17.01} & 3.61 & 16.44 & \textbf{1.92} & -3.4\% & -46.9\% \\
 & LLM-as-a-Judge & \textbf{78.85} & 9.13 & 72.55 & \textbf{4.28} & -8.0\% & -53.1\% \\
\midrule 

\multirow{2}{*}{Math Calculation} & Rouge-L & 18.69 & 3.71 & \textbf{19.64} & \textbf{2.32} & +5.1\% & -37.3\% \\
 & LLM-as-a-Judge & 75.70 & 13.08 & \textbf{78.48} & \textbf{10.09} & +3.7\% & -22.8\% \\
\midrule 

\multirow{2}{*}{Format Conversion} & Rouge-L & 18.76 & 3.98 & \textbf{19.09} & \textbf{2.00} & +1.8\% & -49.7\% \\
 & LLM-as-a-Judge & 80.76 & 11.82 & \textbf{84.65} & \textbf{8.41} & +4.8\% & -28.8\% \\
\midrule 

\multirow{2}{*}{Health Medical} & Rouge-L & \textbf{16.66} & 3.18 & 15.96 & \textbf{1.58} & -4.2\% & -50.1\% \\
 & LLM-as-a-Judge & \textbf{83.49} & 8.71 & 81.47 & \textbf{4.33} & -2.4\% & -50.2\% \\
\midrule 

\multirow{2}{*}{Logic Puzzles} & Rouge-L & 15.34 & 3.18 & \textbf{16.43} & \textbf{1.96} & +7.2\% & -38.4\% \\
 & LLM-as-a-Judge & 73.50 & 14.15 & \textbf{80.32} & \textbf{9.53} & +9.3\% & -32.6\% \\

\bottomrule
\end{tabular}%
}
\caption{\label{tab:main_results} \textbf{Performance and Stability Comparison.} Comparison between the Base prompts and our Improved prompts (incorporating identified robust patterns) across six diverse tasks. Results are averaged over 2,000 samples. \textbf{Mean} denotes average performance (higher is better), and \textbf{Std} denotes standard deviation (lower indicates better stability). The best results are highlighted in \textbf{bold}.}
\end{table*}
\subsubsection{Quantitative Analysis}
The results of this intervention are summarized in Table \ref{tab:main_results}. The data provides strong empirical evidence that our pattern injection strategy significantly mitigates prompt sensitivity across six diverse task domains.

The most striking empirical finding is the universal decline in Standard Deviation across all evaluated tasks and metrics, underscoring a significant gain in stability. Specifically, in Code Writing, which is a domain highly sensitive to syntactic precision, the Improved Prompts achieved a remarkable $40.7\%$ reduction in Rouge-L Std and a $40.4\%$ reduction in LLM-as-a-Judge Std. This result confirms that introducing ``Explicit Action Directives'' effectively constrains the code generation space, preventing the model from drifting into non-functional stylistic variations. The LLM-as-a-Judge metric showed particularly dramatic improvement, with both Mean Score increasing by $7.9\%$ and Std decreasing from $12.96$ to $7.73$, demonstrating enhanced quality alongside stability.

Similar trends were observed in Math Calculation where Rouge-L Std decreased by $37.3\%$ and LLM-as-a-Judge Std by $22.8\%$, demonstrating that ``Domain Anchoring'' successfully ensures model adherence to standard mathematical notations and reasoning steps. This pattern extends to other technical domains such as Format Conversion ($49.7\%$ Rouge-L Std reduction) and Logic Puzzles ($38.4\%$ Rouge-L Std reduction), confirming the broad applicability of our approach.

Beyond stability, our results refute the common concern that enforcing robustness must come at the expense of performance. Instead, constraining the interpretation space frequently enhances quality. Across the six tasks, four domains exhibited simultaneous improvements in both Mean performance and stability. Logic Puzzles achieved the most dramatic Mean Score gains ($+7.2\%$ Rouge-L, $+9.3\%$ LLM-as-a-Judge), while Code Writing and Format Conversion also showed consistent positive lifts across both metrics.

A nuanced trade-off appears in the open-ended domains of Creative Writing and Health Medical, where we observed slight reductions in Mean Scores across both subjective and objective metrics (Creative Writing: $-3.4\%$ Rouge-L, $-8.0\%$ LLM-Judge; Health Medical: $-4.2\%$ Rouge-L, $-2.4\%$ LLM-Judge). We attribute this phenomenon to the \textit{constriction of the solution space}. In these open-ended tasks, unconstrained prompts allow the model to traverse a vast execution topology, occasionally reaching high-scoring local maxima through high variance (``creative but unstable'' outputs). Our pattern injection imposes necessary structural constraints that prune these diverse but unpredictable execution paths. While this regularization slightly lowers the ceiling for stylistic diversity—reflected in the dipped mean scores—it massively raises the floor for reliability. This is confirmed by drastic reductions in variance across both metrics (Creative Writing: $46.9\%$ Rouge-L Std, $53.1\%$ LLM-Judge Std; Health Medical: $50.1\%$ Rouge-L Std, $50.2\%$ LLM-Judge Std). For safety-critical domains like Health Medical, exchanging minor stylistic fluidity for predictable, hallucination-resistant stability represents a crucial optimization for production readiness.

\subsection{Case Study}
\begin{figure}
    \centering
    \includegraphics[width=\linewidth]{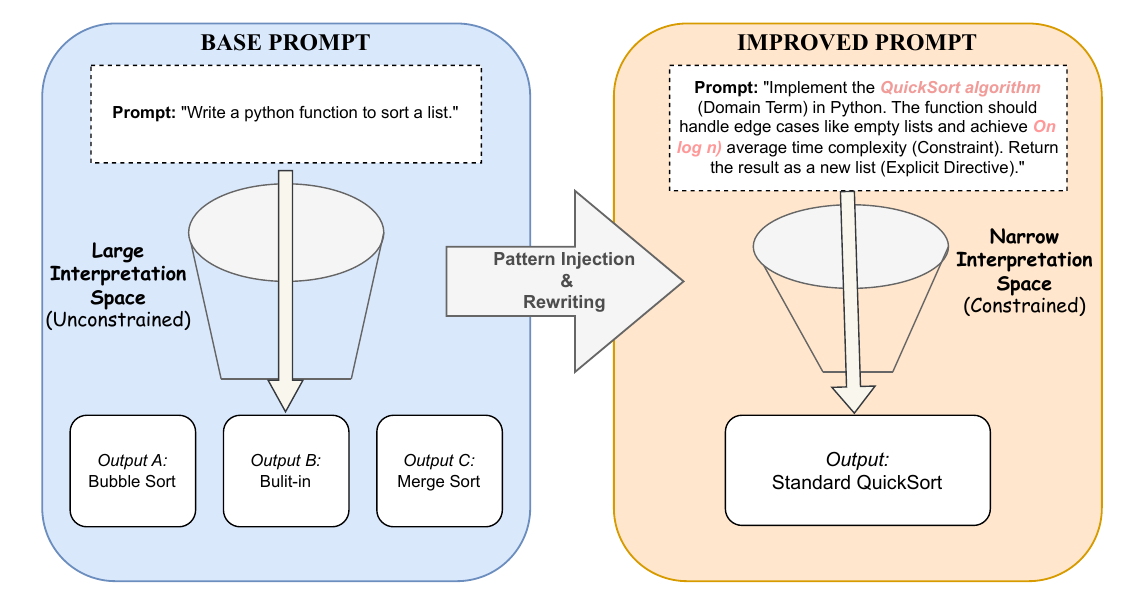}
    \caption{\textbf{Case Study of Prompt Robustness.} The left is the base prompt, while the right is our improved prompt.}
    \label{fig:case_study}
\end{figure}

Figure \ref{fig:case_study} visualizes the mechanistic impact of pattern injection in Code Writing. The Base Prompt (``Write a python function to sort a list'') leaves the \textit{Interpretation Space} unconstrained, resulting in stochastic divergence ranging from built-in methods to inefficient Bubble Sorts. Conversely, the Improved Prompt narrows this space via \textit{Domain Terms} (e.g., ``QuickSort'') and \textit{Explicit Directives}. Acting as locking mechanisms, these features eliminate ambiguity and force reasoning to converge on a standardized path, demonstrating that robustness is achieved by rigorously defining output boundaries rather than suppressing creativity.

\subsection{Practical Implications for Prompt Engineering}
Our findings offer concrete, actionable guidance for designing robust prompts by (1) incorporating domain-specific terminology to clearly signal the task context instead of relying on generic language, and (2) providing explicit operational instructions that specify concrete actions, such as analyze'' or compare,'' along with clear formats and constraints rather than vague directives. Furthermore, (3) combining domain anchoring with execution specification is essential, as the most effective prompts demonstrate a synergy of these patterns to unambiguously define both \textit{what} is relevant and \textit{how} the task should be performed. Crucially, these guidelines are derived not from anecdotal intuition but from the systematic statistical analysis of token patterns established in this study, providing a solid empirical foundation for prompt engineering best practices.

    
    

\section{Conclusion}
This work addresses the challenge of prompt sensitivity in large language models by advancing from anecdotal observations to a systematic, mechanistic analysis. Based on a massive-scale investigation of 132,000 systematic prompt variants, we identify a fundamental scaling law of prompt performance stability, showing that high-performing prompts are intrinsically associated with low output variance. From a mechanistic perspective, we uncover two linguistic anchors underlying this robustness: domain-specific terminology, which delineates semantic boundaries, and explicit action directives, which constrain reasoning trajectories. We operationalize these findings by designing an automated Prompt-Refining Agent. Experimental deployment confirms that by strictly adhering to Pattern Injection, our agent reduces performance variance by up to 40.7\% in code generation tasks. Overall, this work bridges empirical prompt heuristics and principled prompt engineering, demonstrating that robustness emerges from precisely specifying the model’s interpretation space to minimize ambiguity.

\section*{Acknowledgments}
This work is supported partly by Guangdong Provincial Key Lab of Integrated Communication, Sensing and Computation for Ubiquitous Internet of Things (No.2023B1212010007), China NSFC Grant (No.62472366), 111 Center (No.D25008), the Project of DEGP 
(No.2024GCZX003, 2023KCXTD042), Shenzhen Science and Technology Foundation (ZDSYS20190902092853047),
the China NSFC Grant (No. 62372307, No. U2001207),
Guangdong NSF (No. 2024A1515011691), 
Shenzhen Science and Technology Program (No. RCYX20231211090129039),
Shenzhen Science and Technology Foundation (No. JCYJ20230808105906014).

\section*{Limitations}
While our findings offer significant insights, we acknowledge several limitations that chart the course for future research. First, our experiments primarily relied on the Qwen and Gemini model families; while representative of state-of-the-art architectures, the identified scaling laws and token patterns may exhibit different characteristics in smaller-scale models or distinct architectures such as Mixture-of-Experts. Second, the analysis focused on English prompts within the WizardLM benchmark covering code, math, and creative writing, so the generalizability of ``Domain Anchoring'' patterns to low-resource languages or highly abstract reasoning tasks remains to be verified. Finally, we employed automated evaluation to scale our experiments, and although we implemented a multi-dimensional scoring rubric to mitigate bias, automated judges may still possess inherent preferences that differ from human evaluation, particularly in subjective domains like creative writing.
\bibliography{custom}

\appendix

\section{Appendix}
\subsection{Rewrite Prompt}\label{sec:rewrite_prompt}
\begin{figure*}[t]
  \centering
  \includegraphics[width=0.8\linewidth]{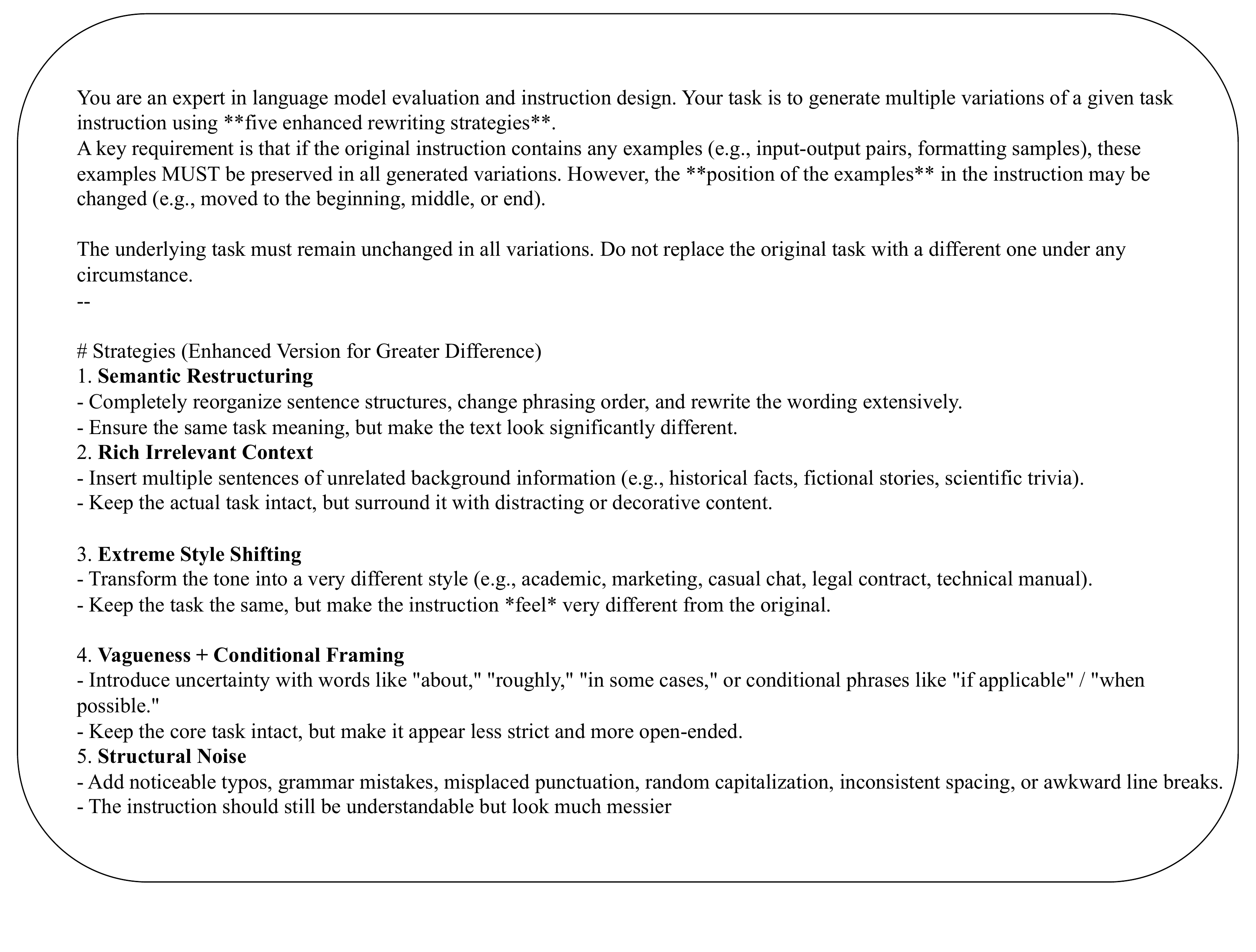}
  \vspace{0.5em}
  \includegraphics[width=0.8\linewidth]{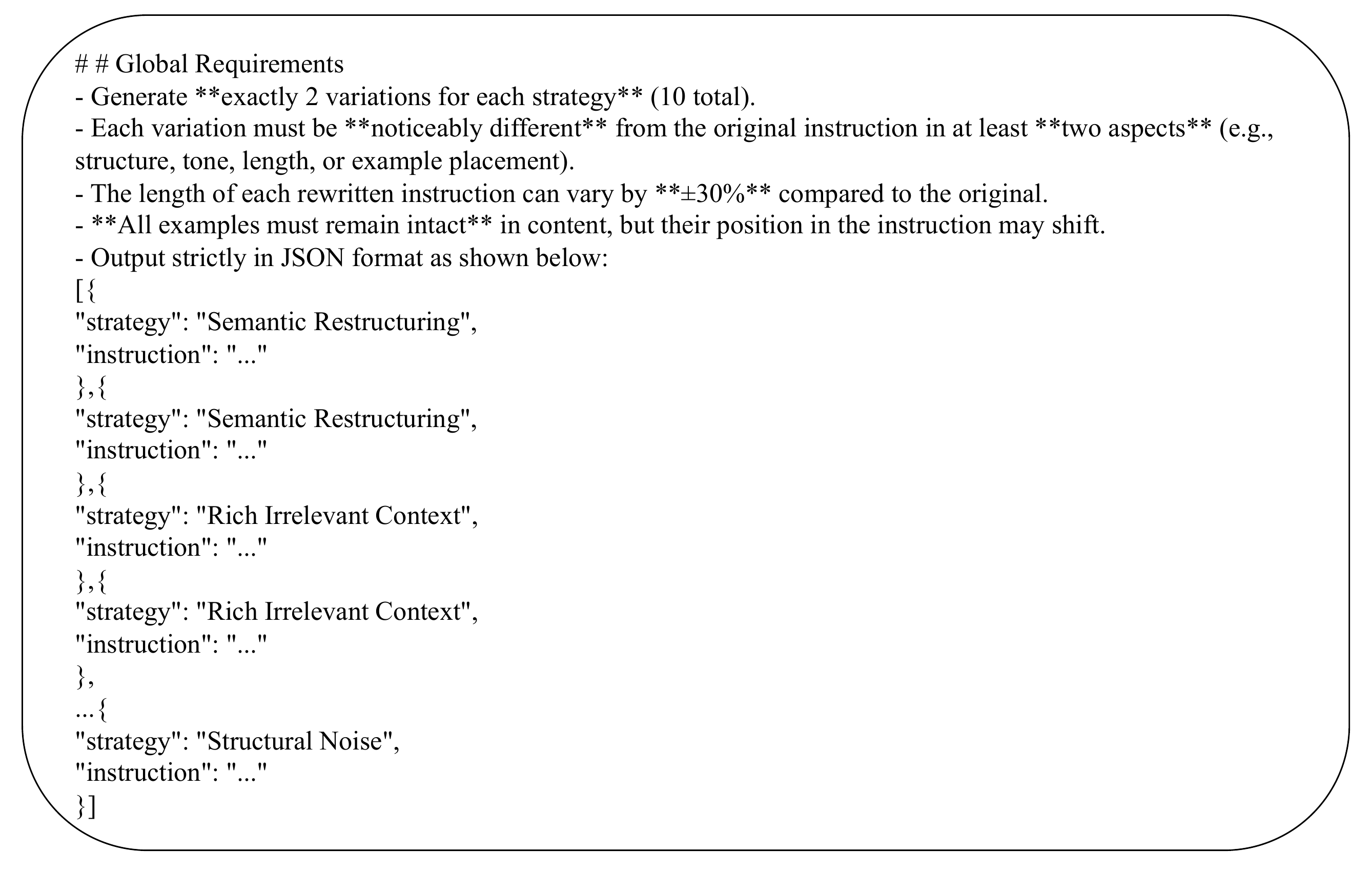}
  \caption{The rewrite prompt to define five strategies.}
  \label{fig:two_images_vertical}
\end{figure*}

To construct the ``Sensitivity Manifold'' of 132,000 prompt variants, we employed Gemini-2.5-flash as the rewriting engine. The model was instructed to generate variations across five orthogonal strategies while strictly preserving the core task semantics and any few-shot examples contained in the original instruction. The full system prompt used for this generation process is provided in Figure \ref{fig:two_images_vertical}.

\subsection{Evaluation Prompt}
\label{sec:eval_prompt}
\begin{figure*}[t]
  \centering
  \includegraphics[width=0.8\linewidth]{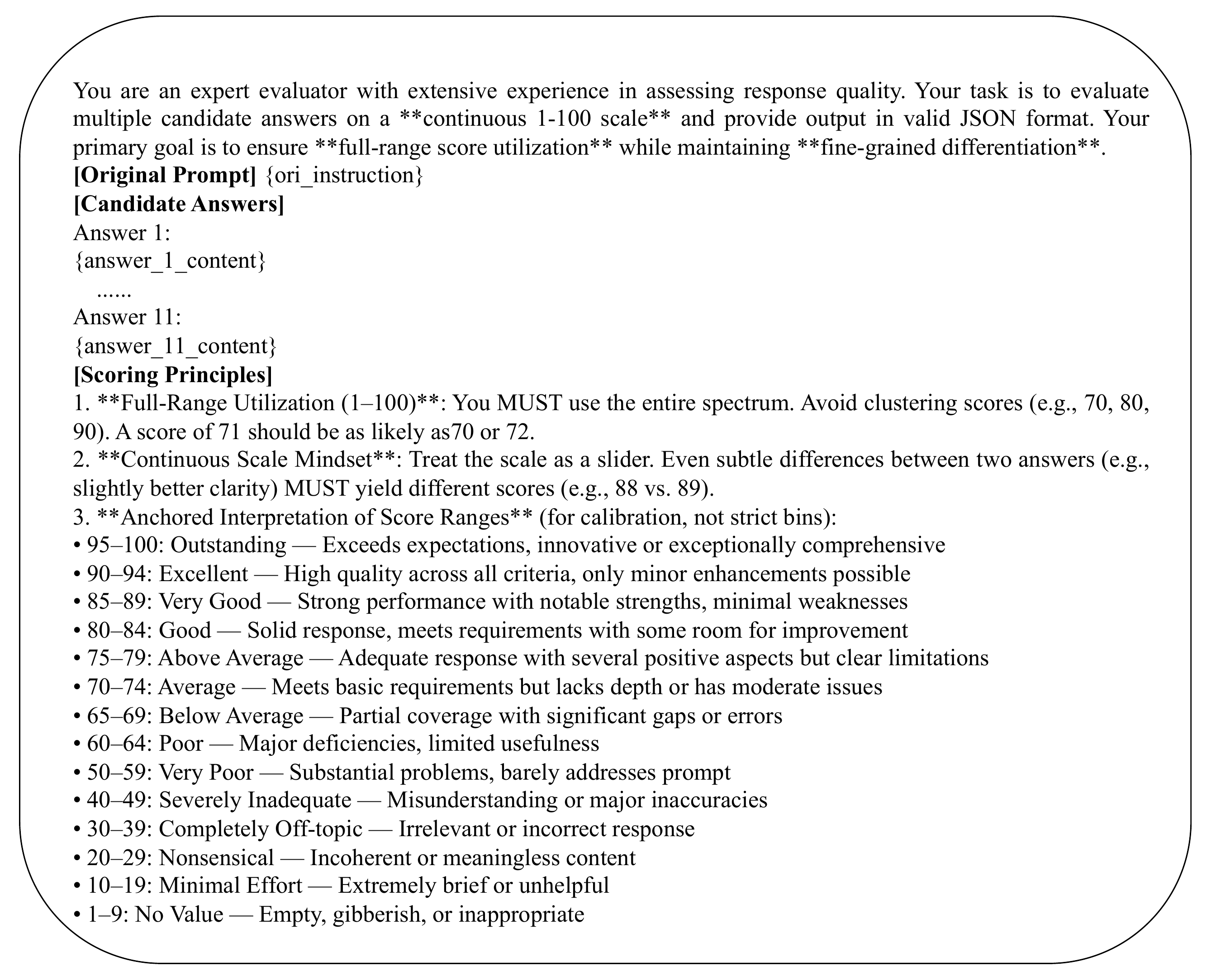}
  \vspace{0.5em}
  \includegraphics[width=0.8\linewidth]{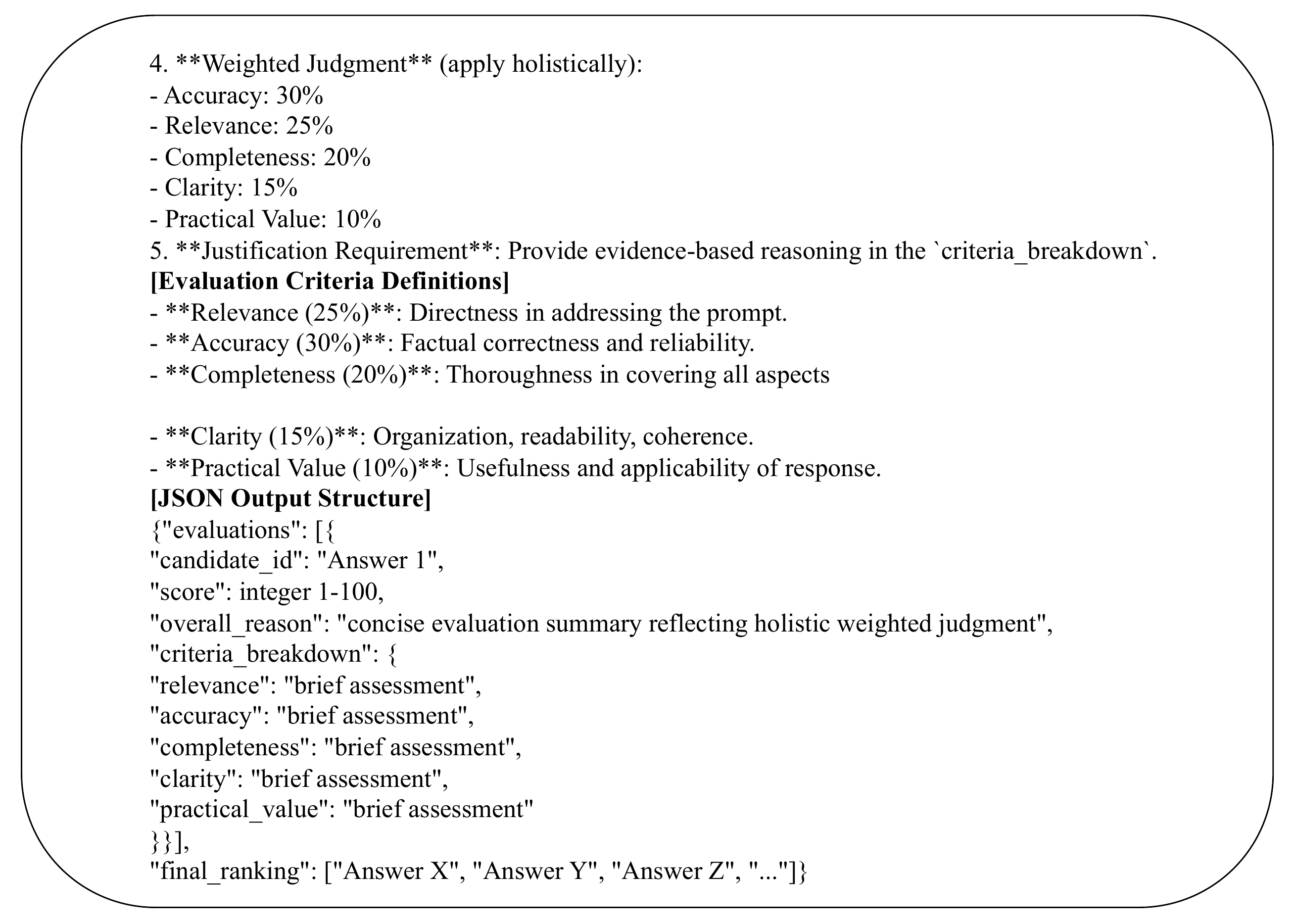}
  \caption{The evaluation prompt of fine-grained quality variations.}
  \label{fig:eval}
\end{figure*}

For the automated quality assessment, we utilize Grok-4-fast as the evaluator. To mitigate the ``clustering effect'' often observed in LLM judges (where scores cluster around 7-10 on a 10-point scale), we designed a continuous 1-100 scoring mechanism with weighted dimensions. The full evaluation prompt is detailed in Fig.~\ref{fig:eval}.

\subsection{Qualitative Comparison of Prompt Variants}
\label{app:prompt_comparison}

Below, we present six representative examples from our dataset, contrasting the original instructions with the robust versions generated by our Prompt-Refining Agent.

\begin{figure*}[h]
    \centering
    \begin{promptbox}{Example 1: Logic Puzzles (Trace Interpretation)}
        \originalprompt{
How can the traveler determine the correct number of open hands after the 2021st gesture based on the pattern provided by the villagers? \\
Here is a possible solution in Ruby: [...] \\
Can you explain how this Ruby code works to solve the puzzle posed by the villagers?
        }
        \improvedprompt{
As an \textbf{expert in Ruby programming and algorithmic sequence computation}, explain how the following Ruby code implements an iterative solution... \\
Decompose your explanation into these \textbf{explicit steps}: \\
1. Initialize variables and describe the loop structure... \\
2. Trace the conditional branches... \\
3. Analyze the final output computation... \\
Output format: Numbered steps with inline code references and a brief trace...
        }
    \end{promptbox}
    \caption{Comparison for Logic Puzzles. The robust prompt anchors the domain to ``algorithmic sequence computation'' and provides a structural template for explanation.}
    \label{fig:ex_logic}
\end{figure*}

\begin{figure*}[h]
    \centering
    \begin{promptbox}{Example 2: Health \& Medical (Safety Guardrails)}
        \originalprompt{
We have some data about a health condition. Summarize the data in a few sentences. \\
Felicity has anxiety. She often experiences physical symptoms, such as a racing heart, shortness of breath, and shakes. [...]
        }
        \improvedprompt{
In the \textbf{domain of clinical psychology}, synthesize a concise summary... incorporating key concepts such as somatic manifestations... \\
Output a 3-4 sentence paragraph: \\
1. Introduce the primary condition... \\
2. Describe cognitive and fear-related aspects. \\
3. Conclude with the affective experience. \\
\textbf{Do not add diagnoses, interpretations, or external information}...
        }
    \end{promptbox}
    \caption{Comparison for Health Medical. The robust prompt enforces strict safety boundaries and structured synthesis.}
    \label{fig:ex_medical}
\end{figure*}

\begin{figure*}[h]
    \centering
    \begin{promptbox}{Example 3: Format Conversion (Grounding Simulations)}
        \originalprompt{
How can I use PHP to parse a JSON object [...] and then retrieve the information for each item to \textbf{take a photograph} of each item...
        }
        \improvedprompt{
As a PHP developer... refactor the provided PHP code... to generate and ``save'' images with descriptive filenames. \\
Key requirements: ... \\
- \textbf{Simulate image capture} and save by using \texttt{file\_put\_contents()} to write a placeholder byte string... \\
- Do not implement actual photography; focus solely on filename generation and file I/O.
        }
    \end{promptbox}
    \caption{Comparison for Format Conversion. The robust prompt translates abstract/impossible requests into concrete engineering simulations.}
    \label{fig:ex_format}
\end{figure*}

\begin{figure*}[h]
    \centering
    \begin{promptbox}{Example 4: Math Calculation (Algorithmic Constraints)}
        \originalprompt{
Can you solve this mountain peak puzzle? [...] the code should not use any built-in Python functions to sort the data or find the maximum value.
        }
        \improvedprompt{
As a Python developer specializing in \textbf{algorithmic constraints and manual data traversal}... \\
\textbf{Procedural Directives:} \\
1. Define the mountain\_peaks list... \\
2. Initialize variables: max\_elevation = -1... \\
3. Iterate through the list using a for loop...
        }
    \end{promptbox}
    \caption{Comparison for Math Calculation. The robust prompt enforces constraints by defining the exact implementation procedure.}
    \label{fig:ex_math}
\end{figure*}

\begin{figure*}[h]
    \centering
    \begin{promptbox}{Example 5: Creative Writing (Structured Generation)}
        \originalprompt{
How can I use C\# code to generate a random plot twist for the given short story using the following Markdown syntax? [...]
        }
        \improvedprompt{
As a C\# developer specializing in \textbf{procedural content generation}... \\
Requirements: \\
1. Initialize a Random instance... \\
2. Define exactly 3 distinct plot twists... \\
3. Output the selected twist... formatted to insert seamlessly... \\
\textbf{Do not alter the original story}; append only the twist.
        }
    \end{promptbox}
    \caption{Comparison for Creative Writing. The robust prompt isolates the generation scope to prevent content drift.}
    \label{fig:ex_creative}
\end{figure*}

\begin{figure*}[h]
    \centering
    \begin{promptbox}{Example 6: Code Writing (Refactoring Stability)}
        \originalprompt{
How can I use Ruby to write a web scraping script... Please let me know if there's anything I can add or modify to improve the script.
        }
        \improvedprompt{
As a Ruby web scraping specialist, \textbf{refactor} the following incomplete Ruby script... to robustly extract prices... \\
Ensure the script handles exceptions... cleans extracted text... \\
\textbf{Do not add new features} like authentication or multi-threading; preserve the core array output.
        }
    \end{promptbox}
    \caption{Comparison for Code Writing. The robust prompt focuses the model on engineering robustness rather than open-ended feature expansion.}
    \label{fig:ex_code}
\end{figure*}

\end{document}